\documentclass{article}

\usepackage{PRIMEarxiv}
\usepackage[labelfont=bf]{caption}
\usepackage[utf8]{inputenc} 
\usepackage[T1]{fontenc}    
\usepackage{hyperref}       
\usepackage{url}            
\usepackage{booktabs}       
\usepackage{amsfonts}       
\usepackage{nicefrac}       
\usepackage{microtype}      
\usepackage{fancyhdr}       
\usepackage{graphicx}       

\usepackage{amsmath,amssymb,amsfonts}
\usepackage{algorithmic}
\usepackage{graphicx}
\usepackage{textcomp}
\usepackage{xcolor}
\usepackage[exponent-product=\cdot]{siunitx}
\usepackage{multirow}
\usepackage{tabularray}
\usepackage{supertabular}
\usepackage{amsmath}
\usepackage{makecell}
\usepackage{float}
\usepackage{xurl}
\usepackage{adjustbox}
\usepackage{orcidlink}

\newcommand{\quoted}[1]{``#1''} 

\DeclareFontFamily{U}{mathx}{\hyphenchar\font45}
\DeclareFontShape{U}{mathx}{m}{n}{<-> mathx10}{}
\DeclareSymbolFont{mathx}{U}{mathx}{m}{n}
\DeclareMathAccent{\widebar}{0}{mathx}{"73}

\graphicspath{{Figs/}}     

\title{Unsupervised Novelty Detection Methods
Benchmarking with Wavelet Decomposition
\thanks{This document is a preprint as at September 10, 2024. The paper has been accepted for publication as proceeding at the \textit{8th International Conference on System Reliability and Safety. Sicily, Italy - November 20-22, 2024} \url{https://www.icsrs.org/index.html}}
}
\author{
  Ariel Priarone \orcidlink{0009-0006-2741-8054} $^\dag$, Umberto Albertin \orcidlink{0000-0002-9328-2590} $^\ddag$, Carlo Cena \orcidlink{0000-0002-7802-5625} $^\ddag$, Mauro Martini \orcidlink{0000-0002-6204-3845}, Marcello Chiaberge \orcidlink{0000-0002-1921-0126}\\
  Department of Electronics and Telecommunications \\
  Politecnico di Torino \\
  Turin, Italy\\
  \texttt{\{ariel.priarone, umberto.albertin, carlo.cena,}\\
  \texttt{mauro.martini, marcello.chiaberge\}@polito.it} \\
  $^\dag$ Corresponding author \\
  $^\ddag$ This publication is part of the project PNRR-NGEU which has received \\ 
  funding from the MUR – DM 351/2022 and MUR – DM 117/2023.
}

\begin{document}
\maketitle

\begin{abstract}
Novelty detection is a critical task in various engineering fields. 
Numerous approaches to novelty detection rely on supervised or semi-supervised learning, which requires labelled datasets for training. 
However, acquiring labelled data, when feasible, can be expensive and time-consuming. 
For these reasons, unsupervised learning is a powerful alternative that allows performing novelty detection without needing labelled samples.
In this study, numerous unsupervised machine learning algorithms for novelty detection are compared, highlighting their strengths and weaknesses in the context of vibration sensing.
The proposed framework uses a continuous metric, unlike most traditional methods that merely flag anomalous samples without quantifying the degree of anomaly. 
Moreover, a new dataset is gathered from an actuator vibrating at specific frequencies to benchmark the algorithms and evaluate the framework. Novel conditions are introduced by altering the input wave signal. Our findings offer valuable insights into the adaptability and robustness of unsupervised learning techniques for real-world novelty detection applications.
\end{abstract}


\section{Introduction}
\label{sec:introduction}
In a world driven by data, it is essential to identify unexpected events. Novelty detection, i.e. the process of identifying new or unknown data that deviate significantly from the expected or established patterns, is crucial in various fields, as it helps in identifying unusual occurrences that could indicate critical events such as system faults, fraud, or emerging trends.

For example, in manufacturing, novelty detection can be used to spot defects in the production plant \cite{umbertoframework}, while in cybersecurity, it can help detect unusual patterns of behaviour that might indicate a security breach \cite{cyberattack}. 
Lastly, in finance, anomaly detection can be vital for fraud detection, allowing institutions to prevent fraudulent transactions before they cause significant damage. Overall, the capability to recognize novelties enhances decision-making, improves safety, and reduces risks across various domains.

Traditional approaches to novelty detection often rely on statistical methods \cite{schein2006rule, isermann2005model}. 
These methods generally depend on predefined thresholds or models of normal behaviour, which can struggle in their ability to generalize to evolving patterns and unseen anomalies.

Artificial Intelligence (AI) has significantly advanced the field of novelty detection. 
Machine Learning (ML) models can learn from data and identify patterns that would be difficult to specify manually. Among AI approaches, supervised learning algorithms require labelled data to train models that can then predict anomalies. Notable examples are neural networks or support vector machines trained on historical data \cite{supanomdet, ZHU2023119937}. Semi-supervised approaches, which use a combination of labelled and unlabelled data, offer a middle ground, as they leverage the available labelled data to improve the learning process while also incorporating unlabeled data to improve the model's robustness and accuracy \cite{carlopireal, umbertouwb}.

Unsupervised learning techniques, which do not require labelled data, are particularly valuable for novelty detection in situations where obtaining labelled data is impractical or impossible, due to a scarcity of data or to the costs associated with labelling it \cite{carlopireal}. These methods include clustering algorithms, like K-means, autoencoders and density-based approaches \cite{Masud2011, Chalouli2017, ZHANG2018, Zhou2019, Pinedo2020, Gattino2023}. Unsupervised algorithms can discover the underlying structure of the data and identify deviations. This capability makes unsupervised approaches versatile and powerful in changing environments. 

To the best of our knowledge, the current literature lacks a comprehensive study which compares Unsupervised Machine Learning (UML) models able to produce a continuous degradation metric, with several preprocessing algorithms.
We believe such a benchmark could provide valuable information to the novelty detection field, serving as a critical resource for researchers and practitioners.

In this study, we benchmarked various unsupervised AI algorithms for novelty detection on a dataset developed in a laboratory using a shaker to record vibrations at given frequencies, artificially generating novel conditions by changing the input wave signal. 
Our method aims at predicting a continuous metric rather than a binary label, as this approach allows us to evaluate the effectiveness and adaptability of these algorithms in identifying novel patterns in complex data structures. We extract a combination of statistical features and wavelet decomposition coefficients from the raw signal to use them as input for the UML models. 
Moreover, we defined a proper set of metrics to compare the performances of the framework configurations.
Lastly, we analyze the performance of each configuration by recording its inference time to measure the computational effort required. The following sections detail our methodology, experiments, and results.

\section{Related Works}
\label{sec:related_work}
Here we give an overview of previous works on novelty detection, focusing mainly on unsupervised approaches.

The K-means algorithm is used in \cite{Zhou2019,Pinedo2020} to label the degradation states of bearings. In \cite{Zhou2019}, the Intelligent Maintenance Systems (IMS) bearing dataset \cite{IMS_data} is clustered using Traditional Statistical Features (TSF) and Mel-frequency Cepstral Coefficients, and the labelled time series are subsequently used to train a Convolutional Neural Network (CNN) recognition model using raw data, without the need to extract new features. Reference \cite{Pinedo2020} computed TSF and Shannon's entropy to perform clustering on IMS and Case Western Reserve University (CWRU) \cite{CWRU_data} datasets. Pinedo et al. converted the time series into 2D representations, as images, using a CNN, i.e. Alexnet \cite{alexnet}, to classify the degradation states.

Reference \cite{Chalouli2017} proposed a method to quantify the health of bearings and validate it on the IMS dataset. They used time-domain features extracted from the vibration signal and applied a cross-correlation filter to remove the redundant features. Then, the K-means algorithm was used to select only the most relevant features through an optimization procedure aimed at obtaining the most dense and separated clusters. The Self-Organizing Map algorithm is then used to compute a health indicator for the bearing.

A method to efficiently initialize the cluster centres in the K-means algorithm is proposed in \cite{Wan21}. It merges the Ant Colony Optimization algorithm with K-means, and it has been validated by applying a three-layer Wavelet Packet Decomposition (WPD) on the CWRU dataset augmented using a sliding window method to prove the applicability of the method to large datasets.

In \cite{ZHANG2018}, the authors proposed a subset-based deep autoencoder model to learn discriminative features from datasets automatically. This approach has been validated on the CWRU, IMS, and Self-Priming Centrifugal Pump (SPCP) \cite{spcp} bearing vibration datasets.

A case study on the IMS dataset is proposed in \cite{Gattino2023}. The authors leverage the knowledge of the fault frequencies of the bearings to attach labels to the data. The features considered initially are TSF and Redundant Second-Generation Wavelet Packaged Transform coefficients, though the dimensionality of the feature space is reduced by applying Principal Component Analysis (PCA). Lastly, K-means, Support Vector Machine (SVM) and agglomerative clustering are used to perform anomaly detection and to identify the failure modes.

Additionally, a powerful data stream classification approach is proposed in \cite{Masud2011}. Notably, it is not a one-class approach, as it manages more than one \quoted{nominal} class, and deals with the emergence of novel classes during the classification task by adding, to the training set, clusters of new data points with a given level of cohesion.

Reference \cite{umbertoframework} presents a machine learning framework for real-time anomaly detection in sensor data, aiding companies in creating datasets for predictive maintenance. It details an optimized software architecture for efficient novelty detection, validated through a digital model and real-world case studies.

\begin{figure*}[t]
    \centering
    \includegraphics[width = \textwidth]{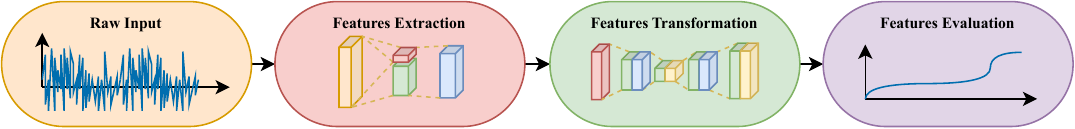}
    \caption{The proposed framework architecture consists of three main blocks. The feature extraction block uses WPD to extract wavelet coefficients. These coefficients, along with statistical measures, form the output of this block. The feature transformation block employs either Autoencoder or PCA to transform the extracted features, enabling the detection of different behaviours in the novelty metric computation. Finally, the feature evaluation block uses six unsupervised machine learning models to compute the novelty metric of the transformed features. 
    }
    \label{fig:architecture}
\end{figure*}

Finally, a computer vision method to detect anomalies in mechanical systems is proposed in \cite{SPYTEK2023109823}. With respect to the previous methods, it has the advantage of evaluating vibrations in multiple points of interest without physical contact with the observed components.

Another model used in the context of novelty detection is the Local Outlier Factor (LOF) \cite{Breunig00}, whose continuous metric depends on the position of the point and the density of known points around it. This concept has been extended to consider also the clustered structure of the data in \cite{HE2003}, which defines the metric Cluster-Based LOF (CBLOF).

Moreover, a simple way of quantifying the normality of data using a distance metric is proposed in \cite{Clifton06}. The authors propose to compute a novelty score, that is the distance of the record from the nearest cluster, normalized by the standard deviation of the distance of the known points in the cluster. This method has been tested on vibration data collected on a jet engine.

Another distance-based method is proposed in \cite{Garcia19}, as it provides a hard classification of the data as normal, extension, or unknown based on the radius of the closest cluster and the position of the point to be evaluated. If the point is outside the decision boundary but within a tolerance, then it is classified as an extension, and the learned model is updated. If the point is outside the tolerance, then it is classified as unknown. Unknown points are kept in a buffer and used to update the model when new classes emerge within the data.

\section{Methodology}
\label{selc:methodology}
\subsection{Novelty detection framework architecture}
\label{subsec:novelty_detection}
The proposed novelty detection framework is unsupervised. The training data used reflects the system's normal behaviour without any labels. Unlike classification algorithms, our framework does not have information about the system's modes. The trained model is then used to detect anomalies in the incoming data, assessing the deviations from the nominal behaviour. The proposed architecture is shown in Fig. \ref{fig:architecture}.

The raw input are the time series collected by an accelerometer sensor. These data are then extracted into a set of key features. These features are then transformed into a latent space that can have lower or higher dimensions. 
Finally, the features are evaluated with several UML models producing a novelty metric (NM) (i.e. the framework's output).
The advantage of this framework is related to provide deviation's severity instead of setting a binary flag (i.e. 0,1) for normal and anomalous behaviours, which is the common approach in anomaly detection tasks. Each block is better explained in the following section.

\subsection{Features Extraction}
\label{subsec:featuresexctract}
The features extracted from the time series are a combination of statistical and WPD coefficients. The statistical features extracted for each signal are: mean, root-mean-square (RMS), peak-to-peak (P2P), standard deviation (STD), skewness and kurtosis. 

\begin{figure}[t]
    \centering
    \includegraphics[]{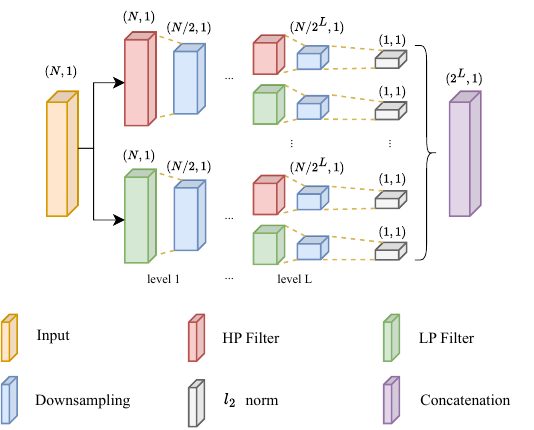}
    \caption{Wavelet Packet Decomposition features extraction architecture. At each level of decomposition, the input signal is split into two sub-band signals, each with a dimension of $N/2$. This process is repeated through $L$ levels, resulting in $2^L$ sub-band signals. For each sub-band signal, the $l_2$ norm is computed, condensing each array into a single value. These values are then aggregated into a final array of dimension $2^L$, forming the complete WPD feature array that encapsulates the characteristics of the original time series data.}
    \label{fig:wavelet}
\end{figure}

The WPD is a tree-based decomposition method able to extract the frequency content of a signal. In this work, a Daubechies wavelet is used.
The decomposition method divides the input data $x \in\mathbb{R}^N$ into two subsets: a set with high-frequency and the other with low-frequency content, each containing $N/2$ elements. This process can be applied an arbitrary number of times (up to $\log_2(N)$) for each subset created. 
The process is repeated recursively until the desired number of levels $L$ is reached.
Hence, the final output is a signal decomposed into $2^L$ vectors in which each vector has $N/2^L$ elements.
Finally, each vector generated by the decomposition is transformed into a single value using the $l_2$ norm. This value assumes the role of a feature. The $2^L$ computed norms form the WPD feature array.
The structure of the decomposition is shown in 
Fig.\ref{fig:wavelet}.

In the end, the WPD feature vector is concatenated with the six statistical features to form the complete features array with dimension $(2^L + 6,1)$ representing the final output of the features extraction block.

After the feature extraction process, the data are normalized, along the whole training dataset, removing the mean and scaling to unit variance in order to use them for the following steps.

At this stage, the extracted features can be directly used to train the models. However, additional processing can be implemented to analyze the different behaviours of the unsupervised model. Specifically, two different Autoencoders and PCA are tested to observe the change in the novelty metric, as explained in the next section.

\subsection{Features Transformation}
\label{subsec:architecture}
In this section, the feature transformation block is explained. Three different algorithms, an undercomplete and an over-complete autoencoder, and PCA are considered.
\subsubsection{Autoencoder}
The first proposed architecture contains an autoencoder (AE), a generative neural network (NN) composed of an encoder that increases or reduces the input dimensionality and a decoder that tries to reconstruct the original input. In this kind of model, the number of neurons in the input and in the output layers must be the same, while the hidden layers can vary in shape and size. The autoencoder is considered undercomplete (AER) if the latent space has a lower dimension than the input and overcomplete (AEA) if the latent space has a higher dimension than the input.
Formally, let $X \in \mathbb{R} ^ n$ represents the input data, $n$ represents the input dimension, the encoder $f_{enc}: \mathbb{R}^{n} \rightarrow \mathbb{R}^{p}$, and the decoder $f_{dec}: \mathbb{R}^{p} \rightarrow \mathbb{R}^{n}$. The autoencoder can be represented as $X_{recon} = f_{dec}(h)$ where $h$ is the latent space $h = f_{enc}(X)$. Hence, the complete form of the eutoencoder can be represented as $X_{recon} = f_{dec}(f_{enc}(X))$. The undercomplete autoencoder has the latent space dimension $p$ lower than the input ($p < n$), while the overcomplete autoencoder has a higher dimension ($p > n$). 
Both NN models have two layers for the encoder and two layers for the decoder.

\begin{figure}[t]
    \centering
    \includegraphics[]{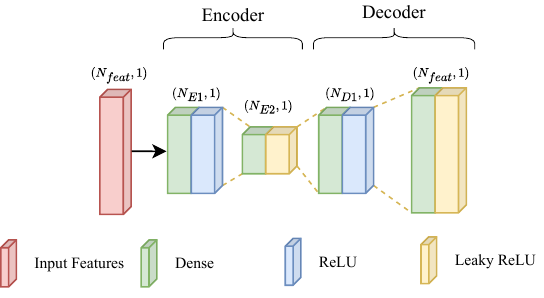}
    \caption{Architecture of the proposed undercomplete autoencoder feature transformation algorithm. The NN model is composed of fully connected layers with the following shape: $N_{E2} < N_{E1}< N_{feat}$ and $N_{E2} < N_{D1} < N_{feat}$. The input is the preprocessed data obtained after the statistical and wavelet transformation.}
    \label{fig:UnderCompleteAE}
\end{figure}

\paragraph{Undercomplete Autoencoder}
The shape of the undercomplete autoencoder is shown in Figure \ref{fig:UnderCompleteAE}.
The input data are the features $n = N_{feat} = 2^L+6$ explained and extracted in Section \ref{subsec:featuresexctract}. The layers have the following dimensions: $N_{E1}$, $N_{E2}$, $N_{D1}$ where $N_{E2} = p$ and $N_{E2} < N_{E1}< N_{feat}$. Then, the decoder transforms the latent space into the original dimension $N_{feat}$, where $N_{E2} < N_{D1} < N_{feat}$. The first layer for both the encoder and decoder uses a ReLU activation function, while the second layer uses a LeakyReLU activation function to enhance the model's robustness and prevent data saturation to zero. The model employs Mean Squared Error (MSE) as the loss function.

\begin{figure}[t]
    \centering
    \includegraphics[]{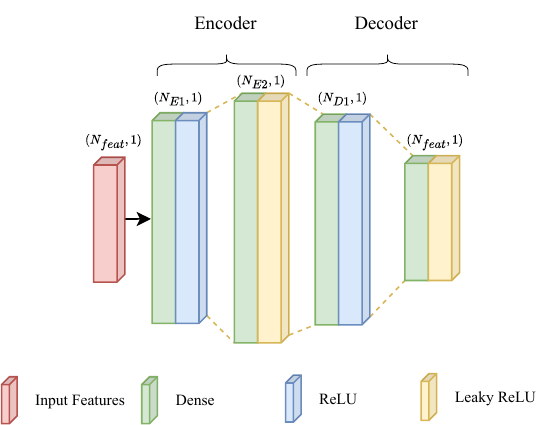}
    \caption{Architecture of the proposed overcomplete autoencoder feature transformation algorithm. The NN model is composed of fully connected layers with the following shape: $N_{feat} < N_{E1} < N_{E2}$ and $N_{feat} < N_{D1} < N_{E2}$.}
    \label{fig:OverCompleteAE}
\end{figure}

\paragraph{Overcomplete Autoencoder}
The same considerations are performed also for the overcomplete autoencoder shown in Figure \ref{fig:OverCompleteAE}.
In this configuration, the input size is increased in the hidden layers of the autoencoder. Formally, the encoder is structured such that $N_{feat} < N_{E1} < N_{E2}$, while the decoder follows $N_{feat} < N_{D1} < N_{E2}$. Both the encoder and decoder employ ReLU activation in the first layer and LeakyReLU activation in the second layer, with MSE as the loss function, similar to the undercomplete autoencoder. 

\subsubsection{Principal Component Analysis}
Principal Component Analysis is employed to transform the features array into a lower dimensional feature space preserving the input variance. The number of features is reduced as a function of the ratio of variance to be preserved.

Formally, let $X \in \mathbb{R}^{n}$ represents the input data, where $n=N_{feat}$. The PCA function is $h = PCA(X)$, where $h$ is the latent space with reduced features $PCA: \mathbb{R}^{N_{feat}} \rightarrow \mathbb{R}^{{N_{PCA}}}$ with $N_{PCA} < {N_{feat}}$.

The three architectures (i.e. AEA, AER, PCA) provide different numbers of features as input to the unsupervised models which produce as output the novelty metric value better explained in the next section.

\subsection{Features Evaluation}
\label{subsec:models}
Here, the six unsupervised models that can be implemented in combination with all the previous feature transformation methods (sec. \ref{subsec:architecture}) are explained in detail. The input of the unsupervised models are the features transformed by the autoencoders or by the PCA. The output of this block is the novelty metric.
When the autoencoders are used, the features are the latent output of the last layer of the encoder, in our case $E_2$.

\subsubsection{K-Means}
The K-Means algorithm is a simple and widely used clustering algorithm. The algorithm requires as input the number of clusters to be created, and the data to be clustered. The algorithm first initializes $k$ centroids and then updates them iteratively, to better fit the data. In this paper, the variant \quoted{K-means++} \cite{Kmeanspp} is used.

The algorithm is trained multiple times with different numbers of clusters to select the best one. The silhouette score is the metric used to choose the number of cluster. Formally, the silhouette score \cite{ROUSSEEUW198753} $s(I)$ of a sample $I$ is a measure of how well the sample fits in its cluster, compared to other clusters, and is defined as:
\begin{eqnarray}
    s(I) & = & \frac{b(I) - a(I)}{\max(a(I), b(I))}\\
    a(I) & = & \frac{1}{|C_i| - 1} \sum_{J \in C_i,I\neq J} d(I, J) \\
    b(I) & = & \min_{k \neq i} \frac{1}{|C_k|} \sum_{J \in C_k} d(I, J)
\end{eqnarray}

where $C_i$ is the cluster assigned to sample $I$, $d(I, J)$ is the distance between samples $I$ and $J$, $C_k$ is another cluster, $|C_i|$ is the number of samples in cluster $C_i$, and $|C_k|$ is the number of samples in cluster $C_k$. 
Hence, $a(I)$ is the average distance between sample $I$ and all other samples in the same cluster, and $b(I)$ is the average distance between sample $I$ and all samples in the nearest cluster. The best number of clusters is selected on the training that achieves the highest average silhouette score across all samples.

Once the model is trained, the centroids are known and a way of measuring how novel the new data are is needed. In this paper, for the k-means models, we propose a Novelty Metric similar to the one in \cite{Clifton06}. The approach involves computing the distance from a new sample to the closest cluster's centroid, comparing this distance with the cluster's radius, and normalizing by this radius (instead of by the standard deviation of the samples in the cluster, as done in \cite{Clifton06}).
The NM is defined as:
\begin{equation}
    \text{NM}_{\text{KMeans}}(I) = \frac{d(I,c_i)-r_i}{r_i}, \quad r_i = \max_{J\in C_i}d(J,c_i)
\end{equation}
where $I$ is the sample to be evaluated, $c_i$ is the closest centroid to $I$ and $r_i$ is the radius of the closest cluster. This normalization accounts for the cluster size while preserving the property of the NM being negative for samples strictly inside the clusters, zero for samples on the boundary of a cluster, and positive for novel samples.

\subsubsection{DBSCAN}
Another clustering algorithm tested is the DBSCAN \cite{dbscan}. It works by dividing the data into \quoted{core} and \quoted{reachable} points, based on the minimum cluster size MinPts and the search radius $\varepsilon$. It has the ability to discard some samples in the training data as noise. In this case, the novelty metric for a new sample $I$ is defined as follows:

\begin{equation}
    \text{NM}_{\text{DBSCAN}}(I) = \dfrac{1}{\varepsilon}\min\limits_{J \in D}d(I,J)
\end{equation}

Where $D$ is the set of samples in the train data that are not marked as noise.

\subsubsection{GMM}

The Gaussian Mixture Model (GMM) is a probabilistic model that assumes that the data is generated from a mixture of several Gaussian distributions. It estimates the parameters of each distribution. In this case, the novelty metric is the likelihood that a given sample belongs to the distribution mixture.

\subsubsection{nuSVM}
\label{nuSVM}
The one-class Support Vector Machine (nuSVM) is a variant of the standard SVM that applies a transformation to the input space in order to separate the data from the origin. New data samples that appear closer to the origin than the training samples are considered outliers.

In this case, we consider the distance of the sample from the separating hyperplane, as proposed by \cite{nuSVM_orig}, as novelty metric $\text{NM}_\text{nuSVM}(i)$.

\subsubsection{IF}
The Isolation Forest (IF) is an ensemble algorithm that trains a set of decision trees to isolate the outliers. The sample is compared with a threshold at each node of the tree, until a leaf is reached. An outlier is expected to reach a leaf in a few steps because it is dissimilar to known samples, while an inlier is expected to reach a leaf after many steps, because it is similar to known samples. We use as novelty metric $\text{NM}_\text{IF}(i)$ the \quoted{anomaly score} defined by the authors \cite{IFOREST_orig}.

\subsubsection{LOF}
The Local Outlier Factor (LOF) is a density-based algorithm that computes the local density of a sample compared to the local density of its neighbours. The algorithm then returns a metric that indicates \quoted{the degree of being an outlier} \cite{Breunig2000}. We consider the result of the algorithm as the $\text{NM}_\text{LOF}(i)$.

\section{Experiments}
\label{sec:exp}

\subsection{Dataset acquisition}
\label{subsec:exp_setup}
A pivotal point of our research is the realization of a dataset to test the Novelty Detection algorithms in a real environment. For this purpose, a shaker (mod. n$^\circ$ K2007E01) and an accelerometer (mod. n$^\circ$ STM32L4R9, an evaluation board of ST microelectronics equipped with an accelerometer sensor) were used to collect a vibration dataset. The evaluation board is mounted on the shaker to gather the dataset with reduced noise and uncertainties.

The shaker is controlled by a PC, reproducing an audio file. Two audio files have been generated, the first reproduces a signal $v_1(t)$ and the second one the signal $v_2(t)$ that we define~as:

\begin{equation}
    v_1(t) = \sum_{i=1}^9 A_i\cdot \sin(2\pi \cdot \alpha_i\cdot t))
\end{equation}
\begin{center}
    $A = \{0.5,0.5,0.5,1,1,1,0.5,0.5,0.5\}$
    $\alpha = \{50,100,150,230,300,440,460,530,600\}$
\end{center}

\begin{equation}
    v_2(t) = \sum_{i=1}^9 B_i\cdot \sin(2\pi \cdot \beta_i\cdot t))
\end{equation}
\begin{center}
    $B = [0.5,0.5,0.5,1,0.2,1,0.5,0.5,2]$
    $\beta = [50,100,150,230,300,440,460,530,600]$
\end{center}

\label{subsec:perfmetrics}

where $A$, $B$ are the weights and $\alpha$, $\beta$ are the frequencies of the harmonics.
The volume setting is used to change the $P2P$ amplitude of the physical signal generated by the PC. 

\begingroup
\renewcommand{\arraystretch}{1.2} 
\setlength{\tabcolsep}{25pt}
\begin{table}[bp]
    \centering
    \caption{Experimental setup for data collection.}
    \label{tab:dataset}
    \begin{tabular}{cccc} 
    \toprule
    \textbf{Set name} & \textbf{Signal Type} & \textbf{P2P [V]} \\
    \hline 
    Set 1 & $v_1(t)$ & 0.25\\
    Set 2 & $v_1(t)$ & 0.50\\
    Set 3 & $v_1(t)$ & 0.75\\
    Set 4 & $v_1(t)$ & 1.00\\
    Set 5 & $v_1(t)$ & 1.25\\
    Set 6 & $v_2(t)$ & 0.50\\
    Set 7 & $v_2(t)$ & 0.75\\
    Set 8 & $v_2(t)$ & 1.00\\
    \bottomrule
    \end{tabular}
\end{table}
\endgroup

Firstly, the signal $v_1$ was fed to the shaker, with five different volume settings, acquiring five sets of data (set 1 - set 5). Every time the volume was changed, the new $P2P$ value was measured with the oscilloscope.  Lastly, this procedure was repeated using the signal $v_2$, to record three more sets (set 6 - set 8). Each of the eight sets of vibration data acquired consists of a record that lasts 206 seconds and has been split into 1-second chunks to generate 206 records. The setup used to collect the eight sets of vibration data is resumed in Table~\ref{tab:dataset}.

The highest frequency chosen is $600$Hz, hence the sampling frequency of the evaluation board is set to $f_s=1666$Hz respecting the Nyquist–Shannon sampling theorem.

\subsection{Features extraction and normalization}
The procedure described in Section \ref{subsec:featuresexctract} is applied to the experimental data collected. The depth level $L$ of the WPD decomposition tree is chosen to be 6, obtaining a WPD feature array with 64 elements. The 6 statistical features are also computed, yielding a final feature array with 70 elements.

\subsection{Models comparison}
The performances of each architecture need to be evaluated in order to perform the benchmark analysis for each architecture. Four different performance metrics are computed for each framework.

\subsubsection{Variance} The variance of the novelty metric is evaluated using the remaining part of the nominal data not used for the training. A good model should produce a stable novelty metric when the data represent the same normal conditions. High variance means that the model is either too sensitive to noise or cannot generalize normal behaviour well. Formally, this metric is defined as:
\begin{equation}
\label{eq:variance}
    Variance = \frac{1}{n} \sum_{i=1}^{n} \left( \text{NM}(i) - \widebar{\text{NM}} \right)^2
\end{equation}
where $\text{NM}(i)$ is the novelty metric of the $i$-th normal sample, $\widebar{\text{NM}}$ is the mean of the novelty metric along all the normal samples, and $n$ is the number of samples.

\subsubsection{Reactivity}
This metric is used to compare different models under the same conditions. The reactivity of the model is defined as the difference between the mean nominal and worst novelty metric. High reactivity indicates that the model can better distinguish between nominal and novelty samples. Formally, the reactivity is defined as: 
\begin{equation}
\label{eq:react}
    \text{Reactivity} = \frac{1}{n} \sum_{i=1}^{n} \text{NM}(i) - \frac{1}{m} \sum_{j=1}^{m} \text{NM}(j)
    \end{equation}
    where $\text{NM}(i)$ is the novelty metric of the $i$-th nominal sample and $\text{NM}(j)$ is the novelty metric of the $j$-th novel sample. 

\subsubsection{Inference Time}
The inference time to evaluate a single sample (i.e. the time interval used by the UML model to compute the NM). This metric is averaged by evaluating several samples. The inference time is measured using an Intel CPU {i9-12900K}.

\subsubsection{Feature Percentage} The Feature Percentage (FP) indicates the percentage of dimensionality preserved after the feature transformation with respect to the original ones. An FP of 100\% means that all original features are retained. A lower FP is preferable for faster computations and to avoid the challenges posed by high-dimensional data (the curse of dimensionality).

\subsection{Models Optimization}

The algorithms, described in section \ref{subsec:architecture}, are also optimized to select the best-performing one for each UML model. 
Concerning the autoencoder, the optimization process is performed by changing the number of neurons of the encoder $N_{E1}$ which is the same for the decoder $N_{E1} = N_{D1}$, the number of neurons of the latent space of the encoder $N_{E2}$, the learning rate, and the batch size. For the PCA, the only parameter changed during the optimization is the number of latent features  $n_f$. 
The variance is chosen as objective function because, at the time of optimization, it is the only variable available without knowledge of the future of the system (e.g. reactivity).
Minimizing this variance is crucial because it reflects the algorithm's ability to model a nominal condition of the system. 
A lower variance indicates better performance because it suggests that the model is less susceptible to noise or environmental changes during the normal operation of the system. 

The optimization used is a Gaussian process-based Bayesian algorithm.
It iteratively suggests values for the hyperparameters that optimize the objective function, aiming to improve the model's robustness and detection capabilities.

The first 100 samples of set 1 of Table \ref{tab:dataset} are used for training, while the remaining 106 samples of the same set are used to compute the variance metric. Formally:

\begin{center}
    $J = min (\sigma_{100}NM)$
    
    $\sigma_{100} (NM) \rightarrow N_{E1,D1},N_{E2},lr,bs,n_f$
\end{center}

where $J$ is the objective function to minimize, $lr$ is the learning rate, $bs$ is the batch size, and $n_f$ is the number of PCA's latent features. The sampler needs a range of values to optimize in order to reach the minimum variance.
The ranges chosen for each variable are reported in Table \ref{tab:aeoptuna}.

\begingroup
\renewcommand{\arraystretch}{1.2} 

\endgroup

The parameters obtained by applying the optimization process are reported in Table \ref{tab:optunaparam}. 

\begin{table}[tbp]
\caption{Hyperparameters ranges chosen for the models' optimization algorithm}
\label{tab:aeoptuna}
\centering
\begin{tabular}{cccccc}
\toprule
\textbf{\makecell{Transf.\\ Method}} & $\mathbf{N_{E1},N_{D1}}$ & $\mathbf{N_{E2}}$ & $\mathbf{lr}$ & $\mathbf{bs}$ & $\mathbf{n_f}$\\
\hline
AEA & $50-65$ & $10-45$  & $0.01-0.1$ & $32,64$ & -\\
AER & $75-80$ & $85-100$ & $0.01-0.1$ & $32,64$ & -\\
PCA & - & - & - & - & $2-70$\\
\bottomrule
\end{tabular}
\end{table}

\begingroup
\setlength{\tabcolsep}{3pt}
\renewcommand{\arraystretch}{1.2} 
\begin{table}[tbp]
\centering
\caption{Tuned hyperparameters. AEA is the overcomplete autoencoder, AER is the undercomplete autoencoder}
\label{tab:optunaparam}
\begin{tabular}{cccccccccc} 
\toprule
\multirow{2}{*}{\textbf{Model}} & \multicolumn{2}{c}{$\mathbf{N_{E1},N_{D1}}$} & \multicolumn{2}{c}{$\mathbf{N_{E2}}$} & \multicolumn{2}{c}{$\mathbf{lr}$} & \multicolumn{2}{c}{$\mathbf{bs}$} & $\mathbf{n_f}$ \\ 
\cline{2-10}
 & AEA & AER & AEA & AER & AEA & AER & AEA & AER & PCA \\ 
\hline
KMeans & 80 & 61 & 85 & 32 & 0.08 & 0.03 & 64 & 64 & 3 \\
DBSCAN & 75 & 56 & 85 & 10 & 0.20 & 0.08 & 32 & 64 & 2 \\
GMM & 79 & 61 & 85 & 10 & 0.08 & 0.01 & 32 & 64 & 2 \\
nuSVM & 80 & 65 & 93 & 10 & 0.09 & 0.10 & 32 & 64 & 2 \\
IForest & 79 & 50 & 85 & 45 & 0.02 & 0.03 & 64 & 64 & 70 \\
LOF & 79 & 50 & 88 & 10 & 0.03 & 0.01 & 32 & 32 & 3 \\
\bottomrule
\end{tabular}
\end{table}
\endgroup

\subsection{Model evaluation}
\begingroup
\setlength{\tabcolsep}{2pt}
\renewcommand{\arraystretch}{1.3} 
\begin{table*}
    \centering
    \caption{Performance metrics for all the preprocessing and model combinations.
    OF means original features, AER means autoencoder reduced, AEA means autoencoder augmented.
    $n_f$ is the number of features transformed by the autoencoders or PCA. 
    FP is the features percentage with respect to the original dimension.
    Variance defined in the Equation \ref{eq:variance}, and the reactivity defined in Equation \ref{eq:react}.
    Train signal is set 1 of Table \ref{tab:dataset} and test signal is set 5.
    The symbols $\Downarrow$ and $\Uparrow$ indicate if the value should ideally be minimized or maximized, respectively. The best results among all the UML models and features transformation combinations are highlighted with *, for each metric.}
    \label{tab:shak_results}
    \sisetup{detect-weight,mode=text}
    \renewrobustcmd{\bfseries}{\fontseries{b}\selectfont}
    \renewrobustcmd{\boldmath}{}
    \newrobustcmd{\B}{\bfseries}
    \begin{adjustbox}{max width=\textwidth}
    \begin{tabular}{
      c
      c
      S[scientific-notation=fixed,table-format=2.0, round-mode = places, round-precision = 0, fixed-exponent=0]
      S[scientific-notation=fixed,table-format=3.2, round-mode = places, round-precision = 2, fixed-exponent=0]
      S[scientific-notation=fixed,table-format=1.5, round-mode = places, round-precision = 4, fixed-exponent=-3]
      S[scientific-notation=fixed,table-format=1.5, round-mode = places, round-precision = 4, fixed-exponent=0]
      S[scientific-notation=fixed,table-format=1.5, round-mode = places, round-precision = 4, fixed-exponent=0]
      S[scientific-notation=fixed,table-format=1.5, round-mode = places, round-precision = 4, fixed-exponent=0]
      S[scientific-notation=fixed,table-format=3.1, round-mode = places, round-precision = 1, fixed-exponent=-6]
      } 
      \toprule
      {\multirow{2}{*}{\textbf{UML model}}}
      &
        {\multirow{2}{*}{\textbf{Transformation}}}
      &
        {\multirow{2}{*}{\textbf{$\mathbf{n_f}$}}}
      &
        {\multirow{2}{*}{\textbf{FP [\%]} $\Downarrow$}}
      &
        {\multirow{2}{*}{\textbf{Variance }[$10^{-3}$] $\Downarrow$}}
      &
        \multicolumn{2}{c}{\textbf{Mean}}
      &
        {\multirow{2}{*}{\textbf{Reactivity }$\Uparrow$}}
      &
        {\multirow{2}{*}{\textbf{Inference Time [}$\mu$\textbf{s] }$\Downarrow$}}
      \\ 
      \cline{6-7}
     & &  &  &  & {\textbf{\textit{Nominal signal }$\Downarrow$}} & {\textbf{\textit{Novelty Signal }$\Uparrow$}} &  &  \\ 
      \hline
\multirow{4}{*}{KMeans} &  OF & 70 & 100.0 & 0.00026796821862546934 & 0.00631025298022692 & \B0.9685670243066067 & \B0.9622567713263798 &  0.00024142449308438318 \\
 & AER & 32 & 45.714285714285715 & 0.00010977015703393284 & 0.007177391028582844 & 0.9351566074476284 & 0.9279792164190456 &  0.00021078502250254346 \\
 & AEA & 85 & 121.42857142857142 & 0.00016153762505665163 & \B0.0013055233914860342 & 0.9611787107095507 & 0.9598731873180647 &  0.00021246238536773389 \\
 & PCA & \B 3 & \B 4.285714285714286 & \B 7.795668354370726e-05 & 0.01903904332395564 & 0.8859691820169525 & 0.8669301386929968 &  \B0.00019312058231071643 \\
\hline
\multirow{4}{*}{DBSCAN} &  OF & 70 & 100.0 & 0.00029936407764120986 & 0.011863781925262917 & \B0.9834854541323141 &\B 0.9716216722070512 &  1.1263936278904366e-05 \\
 & AER & 10 & 14.285714285714285 & 9.571165531785995e-05 &\B 0.0034788697785839017 & 0.9242580562886084 & 0.9207791865100244 &  7.177855807485305e-06 \\
 & AEA & 85 & 121.42857142857142 & 7.755911156776788e-05 & 0.009478196189395651 & 0.9234240870462852 & 0.9139458908568896 &  7.995071901769117e-06 \\
 & PCA & \B2 * & \B2.857142857142857 *& \B1.1127562844358665e-05 & 0.004271087286615494 & 0.909290106600903 & 0.9050190193142875 &  \B6.461066831729803e-06 \\
\hline
\multirow{4}{*}{GMM} &  OF & 70 & 100.0 & \B8.917516103186067e-10 * &\B 4.72103187564289e-06  *&\B 0.9479291821124592 &\B 0.9479244610805836 &  2.2773190709938957e-05 \\
 & AER & 10 & 14.285714285714285 & 8.891762942575812e-07 & 0.0003255883719747679 & 0.8850484168774406 & 0.8847228285054659 &  8.076333539662254e-06 \\
 & AEA & 85 & 121.42857142857142 & 1.224428118343997e-05 & 0.0004747588651318663 & 0.9287239009124388 & 0.928249142047307 &  3.086945634946179e-05 \\
 & PCA & \B2  *& \B2.857142857142857 * & 2.186107332309081e-07 & 0.0005529781894512698 & 0.8321160926415524 & 0.8315631144521012 &  \B5.748877571326743e-06 \\
\hline
\multirow{4}{*}{nuSVM} &  OF & 70 & 100.0 & 0.02316323368773742 & 0.3164999949234617 &\B 1.0  *& 0.6835000050765383 &  2.9675830215503164e-06 \\
 & AER & 10 & 14.285714285714285 & 0.007991048611295305 & 0.06832241388336166 & \B0.9999999999999999 * & 0.9316775861166382 &  \B8.70879440062299e-07  *\\
 & AEA & 93 & 132.85714285714286 & \B0.0021082448459297325 & \B0.057987027623221964 &\B 1.0 * & \B0.9420129723767781 &  1.2518891950895548e-06 \\
 & PCA & \B2  * & \B2.857142857142857  *& 0.002346769585092209 & 0.06303640448302977 & \B1.0  *& 0.9369635955169702 &  1.0495017195821192e-06 \\
\hline
\multirow{4}{*}{IForest} &  OF & 70 & 100.0 & \B0.00455722190040741 &\B 0.11764135059845676 & 0.889347581086465 & \B0.7717062304880082 &  6.0877232689566164e-06 \\
 & AER &\B 45 & \B64.28571428571429 & 0.014981915288497835 & 0.21338245442674392 & 0.9192150161590096 & 0.7058325617322656 &  \B5.258241267066293e-06 \\
 & AEA & 85 & 121.42857142857142 & 0.016206216040787993 & 0.18364287773921348 & 0.9461796223856223 & 0.7625367446464087 &  \B5.328770235803733e-06 \\
 & PCA & 70 & 100.0 & 0.012245213175545725 & 0.25681817744220586 &\B 0.9822948407185857 & 0.7254766632763798 &  5.860803978236158e-06 \\
\hline
\multirow{4}{*}{LOF} &  OF & 70 & 100.0 & 0.0002762363411622762 & 0.0027681405936525807 & \B0.9804187774878576 &\B 0.977650636894205  *&  9.586496751791411e-05 \\
 & AER & 10 & 14.285714285714285 & 8.160021e-05 & 0.0028882856 & 0.9157057 & 0.9128174185752869 &  \B2.6877670042767787e-06 \\
 & AEA & 88 & 125.71428571428571 & 5.2164884e-05 & \B0.0015095939 & 0.944758 & 0.9432483911514282 &  4.504654568491258e-06 \\
 & PCA & \B3 & \B4.285714285714286 & \B6.3705681324758255e-06 & 0.00322729291336856 & 0.8837690364440849 & 0.8805417435307163 &  4.21487250128743e-06 \\

\bottomrule
\end{tabular}
\end{adjustbox}
\end{table*}
\endgroup

\begin{figure}[tbp]
    \centering
    \includegraphics[]{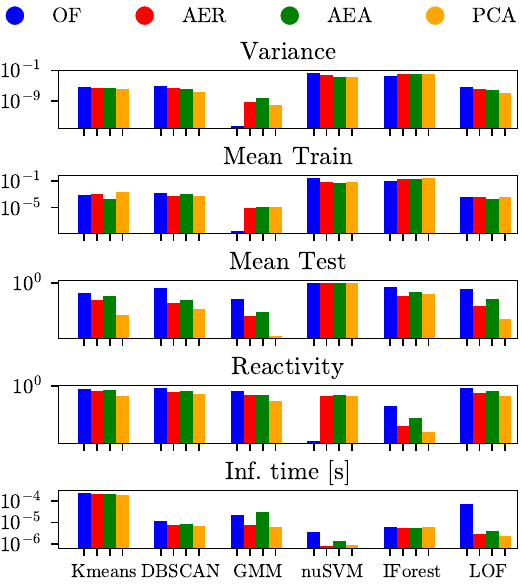}
    \caption{Graphical representation of performance metrics reported on Table~\ref{tab:shak_results}. All the $y$ axes are displayed in a logarithmic scale. AEA is the overcomplete autoencoder, AER is the undercomplete autoencoder and OF is the original features.}
    \label{fig:hist_shak}
\end{figure}

\begin{figure}[tp]
    \centering
    \includegraphics{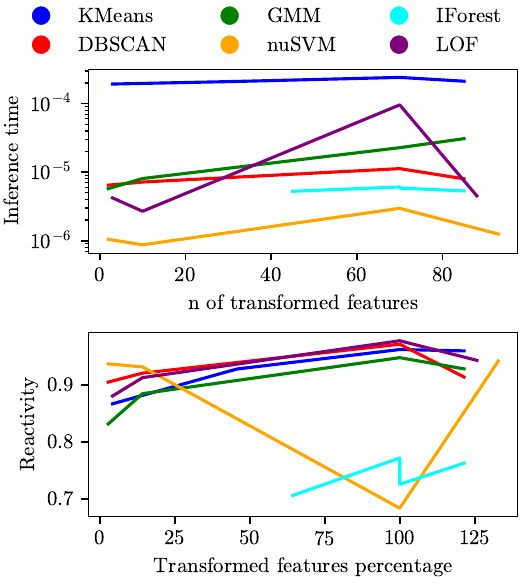}
    \caption{In the upper plot, the correlation between the inference time and the number of features. In the bottom plot, the correlation between the reduced feature dimensionality ratio FP and the reactivity.}
    \label{fig:correlations}
\end{figure}

Performance metrics are computed across all combinations of UML models and feature transformation methods.

The first 100 samples of set 1 are used to train the UML models. Four combinations are benchmarked for each UML model: a test is performed without any features transformation using only the features extracted by the features extraction block (we refer to this scenario with OF, which stands for \quoted{Original Features}). The other three use the undercomplete and overcomplete AE, and the PCA as transformation algorithms.

The remaining 106 samples of set 1 are used to compute the variance and the mean of the nominal signal (i.e. \quoted{mean - nominal signal}. 

The 206 samples of Set 5, are used as a reference test to compute the \quoted{mean - novelty signal} and consequently the reactivity. Set 5 is selected as the reference novelty test because it represents the most dissimilar signal among the sets. 

The novelty metric produced by all models is computed and normalized to a common scale where $\min(\text{NM})=0$ and $\max(\text{NM})=1$ for a fair comparison.

The resulting performance metrics are reported in Table \ref{tab:shak_results}, where the best transformation method for each UML model, for each metric, is highlighted in bold. Additionally, the best results among all the UML models and transformation combinations are highlighted with an asterisk, for each metric.
The same results are also visually depicted in Figure \ref{fig:hist_shak} for clearness.

Looking at Table \ref{tab:shak_results}, we make some considerations about the best UML models and the best transformation methods. 

Regarding the dimensionality reduction of the feature space, the PCA is almost always the transformation that selects the lowest number of output features, with an exception that arises when coupled with the IForest UML model. In this case, the transformation method that further reduces the dimensionality is the AER.

The GMM is the model that produces the least NM variance in nominal conditions (i.e. the NM produced in nominal conditions is almost constant). Concerning the other UML models, KMeans, DBSCAN and LOF experience the least variance when coupled with PCA feature transformation, while IForest produces the most stable NM when used together with original features.

Ideally, when evaluating a nominal signal, the NM should be low. The model that minimizes this value is the GMM used without any feature transformation. Considering the rest of the UML models, there is no feature transformation that appears to minimize this performance metric for most models.

On the other hand, the NM should be high when evaluating a novel signal. The nuSVM model, produces the highest NM regardless the transformations applied, but we observed that this property comes at the cost of a lower capability of generating a continuous NM value, as it tends to saturate to high values even when the signal differs only slightly from the nominal one.

The reactivity appears to be the highest when the OF are used to feed the UML models, except for the nuSVM that has the highest reactivity when combined with AEA. The analysis reveals that the LOF model with the original features shows the highest reactivity, making it the optimal model. The DBSCAN follows behind, demonstrating a similar reactivity value.

The nuSVM is the one that computes the NM the fastest. This is due to the simplicity of the inference procedure, which consists of a distance calculation. Regarding the other UML models, we observe that the fastest response happens when the dimensionality of the feature space is the lowest.

Some observations can be made about the correlation of the performance metrics. Each metric has been plotted against each other to observe emerging patterns. The two most informative plots are shown in Fig. \ref{fig:correlations}. Regarding the correlation between the inference time and the number of features, the general trend suggests that the evaluation process becomes slower as the dimensionality of the feature space increases. In the second plot, the trend suggests that modifying the feature dimensionality ($FP\neq100$) reduces the reactivity of the UML models, except for nuSVM, which exhibits an inverse behaviour.

\begin{figure}[!p]
    \centering
    \includegraphics[]{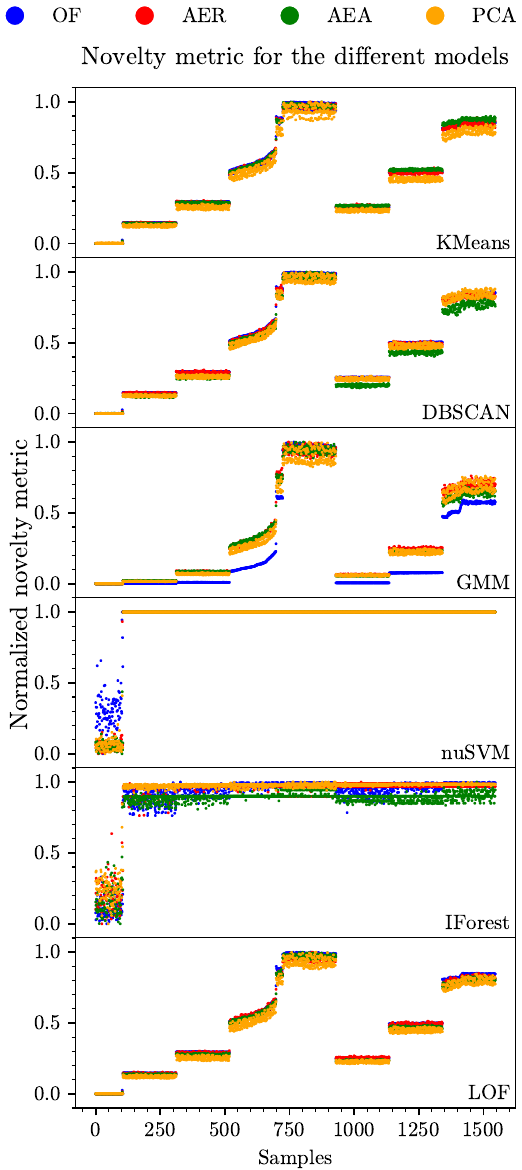}
    \caption{Normalized novelty metric behaviour for all the combinations of UML models and feature transformation technique. In the plots, the steps (discontinuities) in the NM represent the instants in which the framework starts to evaluate a different set of data, with different characteristics. KMeans, DBSCAN and LOF appear less affected by the different feature transformation techniques if compared with the other UML models.
    The nuSVM novelty metric evolution shows a clear saturation for all the samples representative of novel behaviours. Moreover, it fails to produce a steady NM when evaluating samples of known modes. The Iforest model experiences almost the same saturation as the nuSVM. IForest appears also sensitive to the feature transformation used.}
    \label{fig:NM_SHAK}
\end{figure}

For completeness, the novelty metric was computed across the remaining data sets (set 1 - set 8), excluding the portion of set 1 used for training the models. The evolution of the novelty metric along all the sets for each combination is shown in Fig.~\ref{fig:NM_SHAK}. 
It appears evident that KMeans, DBSCAN, GMM and LOF provide a NM that is clearly related to the dissimilarity of the evaluated signal with respect to the nominal condition (at least related to the amplitude increase). On the other hand, as anticipated earlier, nuSVM and IForest exhibits a tendency to saturate the NM to a constant value, with an abrupt gap (almost as a binary flag). This behaviour is in contrast with the scope of our work to provide a continuously changing NM that quantifies the severity of the anomaly. 

The last behaviour that emerges from Fig. \ref{fig:NM_SHAK} is that GMM is quite sensitive to the feature transformation applied. When GMM is evaluating the OF, the NM appears to be lower than when evaluating transformed features.

\subsection{Code availability}
All the functions and scripts used to obtain the results shown in this report will be available online\footnote{\url{https://github.com/PIC4SeR/Unsupervised-Novelty-Detection-Methods-Benchmarking-with-Wavelet-Decomposition.git}}.

\section{Conclusion}
\label{sec:conc}

In this work, we benchmark several unsupervised machine learning models by testing different feature transformation algorithms. We also gather a real vibration dataset to evaluate all frameworks with real-world data, including noise. The results indicate that a continuous Novelty Metric performs better with certain models — specifically KMeans, DBSCAN, GMM, and LOF — by providing an indication of novelty severity. In contrast, models like nuSVM and IF behave as a binary flag. The benchmark highlights the significance of proper feature extraction and transformation algorithms in changing the behaviour of unsupervised models. Additionally, we compute each framework's inference times to assess the complexity of the frameworks in terms of time consumption. Future work will aim to deploy and test each framework on embedded devices to provide an EDGE benchmark. We also plan to test other feature extraction algorithms, such as generative models like RealNVP, to determine if better results can be achieved. 
Finally, we will extend the benchmark tests to other industrial datasets to further investigate the presented study.

\section*{Acknowledgment}
This work has been developed with the contribution of Politecnico di Torino Interdepartmental Centre for Service Robotics PIC4SeR\footnote{\url{www.pic4ser.polito.it}}.

\bibliographystyle{unsrt}  
\bibliography{ref}

\begin{thebibliography}{10}

\bibitem{umbertoframework}
Umberto Albertin, Giuseppe Pedone, Matilde Brossa, Giovanni Squillero, and Marcello Chiaberge.
\newblock A real-time novelty recognition framework based on machine learning for fault detection.
\newblock {\em Algorithms}, 16(2), 2023.

\bibitem{cyberattack}
Adrián Vega, Ignacio Crespo-Martínez, Ángel Guerrero-Higueras, Claudia Álvarez Aparicio, Vicente Matellán, and Camino Fernández.
\newblock Malicious traffic detection on sampled network flow data with novelty-detection-based models.
\newblock {\em Scientific Reports}, 13, 09 2023.

\bibitem{schein2006rule}
Jeffrey Schein, Steven~T Bushby, Natascha~S Castro, and John~M House.
\newblock A rule-based fault detection method for air handling units.
\newblock {\em Energy and buildings}, 38(12):1485--1492, 2006.

\bibitem{isermann2005model}
Rolf Isermann.
\newblock Model-based fault-detection and diagnosis--status and applications.
\newblock {\em Annual Reviews in control}, 29(1):71--85, 2005.

\bibitem{supanomdet}
Jihoon Chung, Bo~Shen, and Zhenyu~James Kong.
\newblock Anomaly detection in additive manufacturing processes using supervised classification with imbalanced sensor data based on generative adversarial network.
\newblock {\em J. Intell. Manuf.}, 35(5):2387–2406, 6 2023.

\bibitem{ZHU2023119937}
Fa~Zhu, Wenjie Zhang, Xingchi Chen, Xizhan Gao, and Ning Ye.
\newblock Large margin distribution multi-class supervised novelty detection.
\newblock {\em Expert Systems with Applications}, 224:119937, 2023.

\bibitem{carlopireal}
Carlo Cena, Umberto Albertin, Mauro Martini, Silvia Bucci, and Marcello Chiaberge.
\newblock {Physics-Informed Real NVP for Satellite Power System Fault Detection}.
\newblock In {\em IEEE/ASME International Conference on Advanced Intelligent Mechatronics (AIM)}, 2024.

\bibitem{umbertouwb}
Umberto Albertin, Alessandro Navone, Mauro Martini, and Marcello Chiaberge.
\newblock Semi-supervised novelty detection for precise ultra-wideband error signal prediction, 2024.

\bibitem{Masud2011}
Mohammad Masud, Jing Gao, Latifur Khan, Jiawei Han, and Bhavani~M. Thuraisingham.
\newblock Classification and novel class detection in concept-drifting data streams under time constraints.
\newblock {\em IEEE Transactions on Knowledge and Data Engineering}, 23(6):859--874, 2011.

\bibitem{Chalouli2017}
Mohammed Chalouli, Nasr-eddine Berrached, and Mouloud Denai.
\newblock Intelligent health monitoring of machine bearings based on feature extraction.
\newblock {\em Journal of Failure Analysis and Prevention}, 17(5):1053--1066, 10 2017.

\bibitem{ZHANG2018}
Yuyan Zhang, Xinyu Li, Liang Gao, and Peigen Li.
\newblock A new subset based deep feature learning method for intelligent fault diagnosis of bearing.
\newblock {\em Expert Systems with Applications}, 110:125--142, 2018.

\bibitem{Zhou2019}
Qicai Zhou, Hehong Shen, Jiong Zhao, Xingchen Liu, and Xiaolei Xiong.
\newblock Degradation state recognition of rolling bearing based on k-means and cnn algorithm.
\newblock {\em Shock and Vibration}, 2019(1):8471732, 2019.

\bibitem{Pinedo2020}
Luis~A. Pinedo-S{\'a}nchez, Diego~A. Mercado-Ravell, and Carlos~A. Carballo-Monsivais.
\newblock Vibration analysis in bearings for failure prevention using cnn.
\newblock {\em Journal of the Brazilian Society of Mechanical Sciences and Engineering}, 42(12):628, 11 2020.

\bibitem{Gattino2023}
Cecilia Gattino, Elia Ottonello, Mario Baggetta, Roberto Razzoli, Jacek Stecki, and Giovanni Berselli.
\newblock Application of ai failure identification techniques in condition monitoring using wavelet analysis.
\newblock {\em The International Journal of Advanced Manufacturing Technology}, 125(9):4013--4026, 04 2023.

\bibitem{IMS_data}
J.~Lee, H.~Qiu, G.~Yu, J.~Lin, and Rexnord~Technical Services.
\newblock Bearing data set.
\newblock IMS, University of Cincinnati, NASA Prognostics Data Repository, NASA Ames Research Center, Moffett Field, CA, 2007.

\bibitem{CWRU_data}
Case Western~Reserve University.
\newblock Bearing data center: Seeded fault test data.
\newblock \url{http://data-acquisition.case.edu/}.
\newblock Accessed: 2024-06-05.

\bibitem{alexnet}
Alex Krizhevsky, Ilya Sutskever, and Geoffrey~E Hinton.
\newblock Imagenet classification with deep convolutional neural networks.
\newblock In F.~Pereira, C.J. Burges, L.~Bottou, and K.Q. Weinberger, editors, {\em Advances in Neural Information Processing Systems}, volume~25. Curran Associates, Inc., 2012.

\bibitem{Wan21}
Lanjun Wan, Gen Zhang, Hongyang Li, and Changyun Li.
\newblock A novel bearing fault diagnosis method using spark-based parallel aco-k-means clustering algorithm.
\newblock {\em IEEE Access}, 9:28753--28768, 2021.

\bibitem{spcp}
Chen Lu, Yang Wang, Minvydas Ragulskis, and Yujie Cheng.
\newblock Fault diagnosis for rotating machinery: A method based on image processing.
\newblock {\em PLOS ONE}, 11(10):1--22, 10 2016.

\bibitem{SPYTEK2023109823}
Jakub Spytek, Adam Machynia, Kajetan Dziedziech, Ziemowit Dworakowski, and Krzysztof Holak.
\newblock Novelty detection approach for the monitoring of structural vibrations using vision-based mean frequency maps.
\newblock {\em Mechanical Systems and Signal Processing}, 185:109823, 2023.

\bibitem{Breunig00}
Markus Breunig, Peer Kröger, Raymond Ng, and Joerg Sander.
\newblock Lof: Identifying density-based local outliers.
\newblock volume~29, pages 93--104, 06 2000.

\bibitem{HE2003}
Zengyou He, Xiaofei Xu, and Shengchun Deng.
\newblock Discovering cluster-based local outliers.
\newblock {\em Pattern Recognition Letters}, 24(9):1641--1650, 2003.

\bibitem{Clifton06}
David~A. Clifton, Peter~R. Bannister, and Lionel Tarassenko.
\newblock Learning shape for jet engine novelty detection.
\newblock In Jun Wang, Zhang Yi, Jacek~M. Zurada, Bao-Liang Lu, and Hujun Yin, editors, {\em Advances in Neural Networks - ISNN 2006}, pages 828--835, Berlin, Heidelberg, 2006. Springer Berlin Heidelberg.

\bibitem{Garcia19}
Kemilly~Dearo Garcia, Mannes Poel, Joost~N. Kok, and Andr{\'e} C. P. L.~F. de~Carvalho.
\newblock Online clustering for novelty detection and concept drift in data streams.
\newblock In {\em Progress in Artificial Intelligence}, pages 448--459, Cham, 2019. Springer International Publishing.

\bibitem{Kmeanspp}
David Arthur and Sergei Vassilvitskii.
\newblock K-means++: The advantages of careful seeding.
\newblock volume~8, pages 1027--1035, 01 2007.

\bibitem{ROUSSEEUW198753}
Peter~J. Rousseeuw.
\newblock Silhouettes: A graphical aid to the interpretation and validation of cluster analysis.
\newblock {\em Journal of Computational and Applied Mathematics}, 20:53--65, 1987.

\bibitem{dbscan}
Martin Ester, Hans-Peter Kriegel, J{\"o}rg Sander, Xiaowei Xu, et~al.
\newblock A density-based algorithm for discovering clusters in large spatial databases with noise.
\newblock In {\em kdd}, volume~96, pages 226--231, 1996.

\bibitem{nuSVM_orig}
Bernhard Schölkopf, Robert Williamson, Alex Smola, John Shawe-Taylor, and John Platt.
\newblock Support vector method for novelty detection.
\newblock volume~12, pages 582--588, 01 1999.

\bibitem{IFOREST_orig}
Fei~Tony Liu, Kai~Ming Ting, and Zhi-Hua Zhou.
\newblock Isolation forest.
\newblock In {\em 2008 eighth ieee international conference on data mining}, pages 413--422. IEEE, 2008.

\bibitem{Breunig2000}
Markus~M Breunig, Hans-Peter Kriegel, Raymond~T Ng, and J{\"o}rg Sander.
\newblock Lof: identifying density-based local outliers.
\newblock In {\em Proceedings of the 2000 ACM SIGMOD international conference on Management of data}, pages 93--104, 2000.

\end{thebibliography}

\end{document}